\documentclass[10pt, journal, web]{article}
\usepackage{times}
\usepackage{caption}
\usepackage[utf8]{inputenc}
\usepackage{utfsym}
\usepackage{multirow}
\usepackage[normalem]{ulem}
\usepackage{amsmath,amssymb,amsfonts}
\usepackage{algorithmic}
\useunder{\uline}{\ul}{}
\usepackage{url}
\def\BibTeX{{\rm B\kern-.05em{\sc i\kern-.025em b}\kern-.08em
    T\kern-.1667em\lower.7ex\hbox{E}\kern-.125emX}}

\usepackage{graphicx}
\usepackage{authblk}

\usepackage{longtable}
\usepackage{pdflscape}
\usepackage[T1]{fontenc}
\usepackage{subfigure}
\usepackage{bbm}
\usepackage[ruled,vlined]{algorithm2e}
\usepackage{pifont}
\usepackage{xcolor}
\usepackage{hyperref}
\usepackage{booktabs}
\hypersetup{
   colorlinks=true,
    linkcolor=blue,
   citecolor=blue,
    urlcolor=blue
}
\usepackage{makecell}
\usepackage{tabularx}
\usepackage{enumitem}

\newcommand{\bvect}{\mbox{\bf b}}

\newcommand{\xvect}{\mbox{\bf x}}

\newcommand{\Dvect}{\mbox{\bf D}}

\newcommand{\Fvect}{\mbox{\bf F}}

\newcommand{\Ivect}{\mbox{\bf I}}

\newcommand{\Lvect}{\mbox{\bf L}}

\newcommand{\Pvect}{\mbox{\bf P}}
\newcommand{\Qvect}{\mbox{\bf Q}}

\newcommand{\Svect}{\mbox{\bf S}}

\newcommand{\Uvect}{\mbox{\bf U}}

\newcommand{\Wvect}{\mbox{\bf W}}
\newcommand{\Xvect}{\mbox{\bf X}}
\newcommand{\Yvect}{\mbox{\bf Y}}
\newcommand{\Zvect}{\mbox{\bf Z}}

\makeatletter
\renewcommand\AB@affilsepx{, \protect\Affilfont}
\makeatother
\usepackage{tabularx}
\providecommand{\keywords}[1]
{
  \small	
  \textbf{\textit{Keywords---}} #1
}

\begin{document}

\title{\textbf{A Re-node Self-training Approach for Deep Graph-based Semi-supervised Classification on Multi-view Image Data}}
\author[1, 2]{Jingjun Bi}
\author[2, 3]{Fadi Dornaika\thanks{Corresponding author}}
\affil[1]{\textit{North China University of Water Resources and Electric Power, NCWU}}
\affil[2]{\textit{University of the Basque Country}}
\affil[3]{\textit{IKERBASQUE}}

\affil[ ]{

\small\texttt{bijingjun@ncwu.edu.cn, fadi.dornaika@ehu.eus}}
\date{}
\maketitle
\begin{abstract}
Recently, graph-based semi-supervised learning and pseudo-labeling have gained attention due to their effectiveness in reducing the need for extensive data annotations. Pseudo-labeling uses predictions from unlabeled data to improve model training, while graph-based methods are characterized by processing data represented as graphs. However, the lack of clear graph structures in images combined with the complexity of multi-view data limits the efficiency of traditional and existing techniques. Moreover, the integration of graph structures in multi-view data is still a challenge. In this paper, we propose Re-node Self-taught Graph-based Semi-supervised Learning for Multi-view Data (RSGSLM). Our method addresses these challenges by (i) combining linear feature transformation and multi-view graph fusion within a Graph Convolutional Network (GCN) framework, (ii) dynamically incorporating pseudo-labels into the GCN’s loss function to improve classification in multi-view data, and (iii) correcting topological imbalances by adjusting the weights of labeled samples near class boundaries. Additionally, (iv) we introduce an unsupervised smoothing loss applicable to all samples. This combination optimizes performance while maintaining computational efficiency. Experimental results on multi-view benchmark image datasets demonstrate that RSGSLM surpasses existing semi-supervised learning approaches in multi-view contexts.
\end{abstract}


 \hspace{10pt}
%%%%%%%%%%%%%%%%%%%%%%%%%%%%%%%%%%%%%%%%%%%%%%%%%%%%%%%%%%%%%%%%%%%%% 

\section{Introduction}

Graph-based methods and pseudo-labeling approaches have recently gained considerable popularity in semi-supervised learning (SSL). Graph-based SSL aims to train models by leveraging both labeled and unlabeled data, while employing a graph structure to capture relationships between data points. Graph learning, particularly through deep graph neural networks \cite{PENG2021401, Thiede2021, Tolstikhin2021, Wu2021}, has become a focal point of research. As data annotation is both costly and time-consuming, many researchers are exploring SSL on graphs. The primary goal in such applications is to infer labels for a large number of unlabeled samples using only a few labeled samples, commonly referred to as node classification. This objective underscores the rising importance of SSL on graphs.

On the other hand, SSL methods based on pseudo-labels use model-predicted labels for unlabeled data as if they were actual labels, thereby increasing the amount of supervision. The core idea is to use early model estimates to predict labels for the unlabeled data, and then incorporate these predictions into the training set to improve the model's generalization capability. Incorporating pseudo-labels refines class boundaries, making the classes more compact. Effective pseudo-label selection ensures that only high-confidence predictions are added to the training set, while noise-reduction techniques maintain the integrity of the training data, thereby boosting overall model performance. Selecting suitable pseudo-labels \cite{rizve2021defense, mukherjee2020uncertainty} and reducing the impact of noise \cite{wang2021self} are key research challenges in this area.

In practice, a significant challenge lies in processing multiple data collections from different sources, i.e., multi-view data \cite{2008An, 2011Co}. In real-world applications, multi-view data often involves different data sources or distinct methods of data representation, such as various data types or techniques that capture the same data from different perspectives. For example, in facial recognition, data can be captured through 2D images, 3D models, and heat maps, each providing unique insights. For image data, features such as Local Binary Patterns, Histograms of Oriented Gradients (HOG), and Color Moments are common representations. Table \ref{datasets8} illustrates several examples of image descriptors.

The advantage of multi-view data lies in its ability to merge information from various perspectives or features, enhancing the effectiveness of data analysis and modeling \cite{2016Consensus, 2021Deep, 2021Self}. 
% Typical strategies for learning from various perspectives include several techniques, such as Co-regularization \cite{2008An}, Co-training \cite{2011Co}, Multiple Kernel Learning \cite{M2011Multiple}, Subspace Learning \cite{2012Multiview}, Margin-Consistency Algorithms \cite{2016Consensus},  and so on \cite{2021Deep, 2021Self}. 
Despite its advantages, multi-view data poses several challenges. First, effectively integrating information from diverse views is difficult due to variations in data distribution and scope, requiring sophisticated fusion methods. Second, ensuring correlation and consistency across views is challenging, as multi-view learning typically assumes some association between perspectives, though noise and inconsistencies can arise. Effective strategies should be developed to manage these issues. Finally, selecting suitable feature representations and learning approaches is crucial, as each view can exhibit distinct representations. A thorough investigation is needed to effectively integrate these representations and leverage their informative potential.

In summary, integrating information from multiple views can alleviate the limitations associated with individual perspectives. However, learning a model from multi-view data remains a complex task due to the high-dimensional nature of image data and the inherent diversity among different perspectives.

Current multi-view semi-supervised learning methods typically face several challenges, including:

\begin{itemize}

% \item Lack of  Explicit Graph Structure: Many methods struggle with datasets that do not have an inherent graph structure, making the task of constructing meaningful graphs more complex.

% \item Graph Construction: Most techniques depend heavily on K-nearest neighbor graphs, leaving room for improvement in designing more suitable and informative graph construction approaches that can enhance the learning capability of the framework.

\item \textcolor{black}{Graph Construction: Many methods struggle with datasets lacking an explicit graph structure, making it challenging to construct meaningful graphs; most techniques rely heavily on K-nearest neighbor graphs, indicating a need for more effective and informative graph construction approaches that could enhance the framework's learning capabilities.} 

\item Fused Feature/Graph: The quality of data across different views can vary significantly. Therefore, it is important to assign appropriate weights to each view during feature or graph fusion to reflect their relative importance.

\item Topological imbalance: Graphs constructed on partially labeled data often suffer from imbalanced graph representation due to the unequal roles of labeled nodes within the graph's topology. This leads to challenges in ensuring that labeled nodes are represented in a structurally balanced way across the graph.

\end{itemize}

Although graph convolutional networks (GCNs) are widely used in graph-based SSL due to their ability to capture discriminative node representations, extending GCNs to handle multi-view data without predefined graphs introduces substantial challenges.
In this paper, we address these problems by presenting efficient graph construction and fusion techniques specifically designed for data with multiple views and without explicit graph structures. Additionally, we propose a seamless scheme that integrates pseudo-labels into the GCN training process through a dynamic loss function. This dynamic pseudo-label loss element offers several advantages:
(i) It avoids the additional learning stages and models typically required by many pseudo-labeling methods.
(ii) It dynamically adjusts the weight of all predictions based on their confidence levels, eliminating the need to set a fixed threshold for pseudo-labels, thus ensuring no pseudo-label information is discarded due to low confidence.
(iii) It mitigates the negative impact of incorrect predictions during early training, reducing the risk of misleading the model with inaccurate pseudo-labels.

Building on these improvements, we propose a Re-node Self-taught Graph-based Semi-supervised Learning for Multi-view data (RSGSLM). 
The main contributions of this study are:

% \item Graph Construction Without K-Nearest Neighbor Dependency: Our approach eliminates the need for ad-hoc K-nearest neighbor graphs by employing an SSL method to generate and adaptively fuse the view graphs.

% \item  Dimensionality Reduction: The dimensionality of the features is significantly reduced by a flexible linear projection technique, which is performed simultaneously with the construction of the view-specific graph, resulting in lower computational costs.

% \item View Contribution Consideration: The method effectively balances the contributions of different views in the graph fusion process, ensuring that each view is weighted appropriately.

\begin{itemize}

\item  \textcolor{black}{Graph Construction and Dimensionality Reduction: Our method eliminates the reliance on K-nearest neighbor graphs by utilizing a semi-supervised learning approach to generate and adaptively fuse view graphs. Concurrently, we apply a flexible projection technique that significantly reduces feature dimensionality, ultimately leading to lower computational costs.}

\item \textcolor{black}{Balanced View Contributions and Topological Imbalance Resolution: This approach ensures that multiple views contribute adaptively to the graph fusion process by effectively assigning appropriate weights to each view. Furthermore, it tackles topological imbalances by adjusting the weights of labeled nodes based on their proximity to class boundaries.}

\item  \textcolor{black}{Integration of Pseudo-labeling: We introduce a novel flexible pseudo-labeling method into the GCN training loss function, enhancing the model's ability to effectively manage unlabeled data.}

\item  \textcolor{black}{Efficiency and Validation: The entire framework operates efficiently with a single GCN, which minimizes both runtime and computational complexity. Our approach demonstrates superior performance across six benchmark multi-view image datasets, as validated by extensive experimentation with ten different methods.}
\end{itemize}

The structure of this paper is as follows: In Section 2, we review related work to contextualize our findings. Section 3 provides essential background information on the methods underpinning our research. In Section 4, we delve into the details of our RSGSLM. Following that, Sections 5-7 outline our experimental setup, analyzes the results, and verify the effectiveness of our pseudo-label loss function through extensive ablation studies. Finally, Section 8 concludes the paper and discusses future research directions.

\section{Related work}

Deep learning, a powerful approach within machine learning, leverages neural networks with multiple layers to detect intricate characteristics and structures in data. This method has demonstrated exceptional performance, particularly in image processing and natural language understanding, surpassing traditional techniques. Nonetheless, data frequently exists not merely as straightforward vectors or tensors, but rather as graphs that represent intricate connections between entities—examples include social networks, biological networks, or transportation systems. These graphs contain rich, untapped information, prompting researchers to apply deep learning techniques to analyze graph data and uncover hidden patterns and insights \cite{zhang2020deep}.

Graph neural networks (GNNs) have surfaced as a fundamental framework for handling data structured as graphs\cite{chen2024survey}. Among the various types of GNNs, Graph Convolutional Networks (GCNs) have gained particular prominence due to their ability to directly handle graph data in semi-supervised learning tasks \cite{kipf2016semi}. Recent advancements in semi-supervised learning with GCNs, such as Adaptive Graph Convolutional Networks \cite{li2018adaptive}, CayleyNet \cite{2018cayleynets}, Diffusion CNN \cite{atwood2016diffusion},  and Dual Graph Convolutional Networks \cite{zhuang2018dual}, have demonstrated strong performance across a variety of domains.

Specialized models have also been developed for multi-view graph structures. For instance, M2GRL \cite{2020M2GRL} enhances recommendation systems by building multi-perspective graphs based on user interactions. In the realm of bioinformatics, MMGCN \cite{Xi2018Multi} improves predictions of miRNA-disease associations through the use of multi-channel attention mechanisms. Additionally, FMCGCN \cite{FMCGCN2025} has demonstrated outstanding performance in the bioinformatics sector. MT-MVGCN \cite{MTMVGCN} excels in social network analysis by extracting insights from complex network data across multiple objectives.
PCConv and its implementation, PCNet \cite{pccon2024}, enhance the graph heat equation to accommodate long-range heterophilic aggregation. They achieve this by employing a dual-filtering mechanism that extracts homophily from heterophilic graphs.

Despite these advances, many types of data, such as image collections, do not naturally form graph structures. A key challenge for graph-based semi-supervised methods is leveraging hidden features and relationships within the data to boost model efficacy. Alternative methods like DSRL \cite{DSRLwang} use sparse regularization to obtain concise latent representations without relying on graphs, improving generalization to unlabeled data. Similarly, AMSSL \cite{AMSSL2021} introduces adaptive fusion techniques, and JCD \cite{JCD} optimizes view-specific classifiers for multi-view SSL. LACK \cite{lack2022semi} dynamically modifies the importance of each view according to labeling data, mitigating the impact of less informative views.
ASFL \cite{ASFL2025} creates a bipartite graph for each view and aligns anchors across views based on sample-anchor relationships, preserving complementarity and consistency. 
\textcolor{black}{CFMSC \cite{CFMSC202337} aims to effectively perform multi-view semi-supervised classification through an adaptive collaborative fusion method that simultaneously integrates multiple feature projections and similarity graphs, thereby enhancing processing efficiency and adaptability.
AMSC \cite{AMSC2024} effectively classifies incomplete multi-view data in semi-supervised scenarios by constructing partial graph matrices to assess relationships among present samples and learning both view-specific and common label matrices to achieve accurate classification for unlabeled data points.
SMFS \cite{SMFS2024} is a technique designed to simultaneously select informative features and learn a unified graph from heterogeneous feature spaces, effectively addressing the challenges posed by unreliable similarity graphs and the diversity of feature projections in the presence of abundant unlabeled data.}

While deep SSL techniques for image data are still developing, models like DSRL and LACK provide valuable strategies. Recent research increasingly targets the development of deep SSL techniques tailored for image data.

As graph-based research has progressed, simpler algorithms like K-Nearest Neighbors (KNN) have been used for graph construction. For example, Co-GCN \cite{coGCN2020} combines co-training with GCNs, using a weighted fusion of graph Laplacians derived from multi-view data obtained via KNN. Similarly, LGCN-FF \cite{LGCNFF2023} integrates graph and feature fusion using adaptive GCNs and a differentiable activation function to enhance graph representation learning. IMvGCN \cite{ImvGCN2023} further bridges GCNs with multi-view learning by integrating reconstruction error with Laplacian embedding and enforcing orthogonality in its transformation matrix. TUNED \cite{TUNED2025} presents a novel MVC framework that seamlessly combines local and global neighborhood structures, improving feature extraction and fusion by capturing cross-view dependencies and reducing conflicts. It features a selective Markov random field (S-MRF) model and a parameterized evidence extractor that dynamically learns and merges evidence from multiple views. While this approach significantly improves the model’s capacity to handle complex and heterogeneous data sources, it employs the fundamental Clustering with Adaptive Neighbors (CAN) method for graph construction.
\textcolor{black}{Both the KNN method, employed in models such as Co-GCN, GCN-FF, and IMvGCN, and the CAN (Correlation Analysis-based Network) method, adopted in the TUNED model, utilize simple unsupervised strategies for graph construction based solely on pairwise similarities or distances. In contrast, the proposed model adopts a more advanced and efficient semi-supervised graph construction strategy, which integrates feature smoothing, label smoothing, label fitting, and regularization—ultimately leading to more accurate graph estimation across views.}

These methods represent significant advances in multi-view graph-based SSL and illustrate different strategies for addressing complex classification tasks. However, deploying deep GCN embedding for SSL on multi-view data without explicit graphs remains under-explored.

\section{Background}
In this section, we will first present the terms used in this study and then describe the GCN model.

\subsection{Notations}

In this study, we use bold uppercase letters to signify matrices and bold lowercase letters to represent vectors. With single view, we use the matrix $\Xvect = [\xvect_1; \ \xvect_2; \ldots; \xvect_n] \in \mathbb{R}^{n \times d}$, which consists of $n$ samples with $d$ features and each row represents one sample. The graph $\Svect \in \mathbb{R}^{n \times n}$ denotes the pairwise similarity between the samples. The diagonal elements of $\Dvect$ are derived by summing the rows in $\Svect$. Additionally, $\Ivect$ represents the identity matrix, and $\Lvect = \Ivect - \Dvect^{-\frac{1}{2}} \Svect \Dvect^{\frac{1}{2}}$ represents the normalized Laplacian matrix.

With multi-view, the data comprises $V$ views, with each view containing $n$ samples and $d_v$ features. The matrix for the $v$-th view is denoted as $\Xvect^v = [\xvect^v_1; \ \xvect^v_2; \ldots; \xvect^v_n] \in \mathbb{R}^{n \times d_v}$. Similarly, for each view, we can construct the corresponding similarity graph $\Svect^v$.

The ground truth label $\Yvect \in \mathbb{B}^{n \times c}$ is constructed using one-hot vectors, where $c$ denotes the number of classes. $\Yvect^p   \in \mathbb{R}^{u  \times c }$  presents the matrix of pseudo labels set to the output predictions of the deep network. In the context of GCN, $\Wvect^{(k)} \in \mathbb{R}^{d_k \times d_{k+1}}$ denotes the learnable layer-specific weight matrix for the $k$-th layer, with $d_0 = d$. The output $\Zvect$ of the GCN is a matrix of size $\mathbb{R}^{n \times c}$. Key notations are described in Table \ref{notation}.

\begin{table}[]
\caption{Key notations summarized in this paper.}
\label{notation}
\centering
\resizebox{1.0\linewidth}{!}{
\begin{tabular}{ll}
\hline
Notation     & Description  \\ \hline
$n$ & Number of data samples\\
$u$ & Number of unlabeled samples\\
$d$ & Dimensionality of data\\
$c$ & Number of classes\\
$V$ & Number of views\\
$\Xvect^v= [\xvect_1; \ \xvect_2; \ldots; \xvect_n]\in \mathbb{R}^{n\times d_v}$ & The data matrix in $v$-th view\\
$d_v$ & Dimensionality of data $\Xvect^v$\\
$\Svect^v \in \mathbb{R} ^{n\times n} $ & Graph similarity matrix of data $\Xvect^v$\\
$\Lvect^v \in \mathbb{R} ^{n\times n} $ & Normalized Laplacian matrix of graph $\Svect^v$\\  
$\Yvect\in \mathbb{B} ^{n\times c}$ & Ground-truth label matrix\\
$\Yvect^p   \in \mathbb{R}^{u  \times c }$ & The matrix of pseudo labels\\
$\Fvect^v  \in \mathbb{R} ^{n\times c}   $ & The soft label prediction of $\Xvect^v$ in linear projection\\

$\Zvect \in \mathbb{R} ^{n\times c}$ & The output of GCN \\
      
$K$ & Number of layers in GCN \\
$\Wvect ^{(k)}\in \mathbb{R} ^{d_k\times d_{k+1}}$, $d_0 = d $ & The learnable weight matrix for the $k$-th layer in GCN\\
$\Svect$ $ \in \mathbb{R} ^{n \times n}$ & Fused  graph matrix \\
$\Lvect \in \mathbb{R} ^{n\times n }$ & Normalized Laplacian matrix of graph $\Svect$ \\ 
$|| \cdot ||$ &   $\ell_2$ norm of a vector or matrix \\

 \hline
\end{tabular}
}
\end{table}

\subsection{Graph convolution network}

Graph Convolution Networks (GCN) \cite{kipf2016semi} use filtered data matrices where the nodes are subjected to graph convolution by multiplying normalized graph representations with feature matrices. Consequently, propagation over $K$ layers in a GCN is formulated as follows:

\begin{equation}\label{xk1} \Xvect^{(k+1)}=\sigma(\Dvect^{-\frac{1}{2}}\Svect\Dvect^{-\frac{1}{2}}\Xvect^{\left(k\right)} \Wvect ^{\left(k\right)}) \end{equation}
here $k = 0, 1,..., K - 1$, and $ \Xvect^{(0) } = \Xvect$. $\sigma(\cdot)$ denotes the non-linear activation function.

The last layer can be defined as: 
\begin{equation}\label{z1}
\Zvect=softmax(\Dvect ^{-\frac{1}{2}} \Svect \Dvect ^{-\frac{1}{2}} \Xvect ^{\left(K-1\right)} \Wvect ^{\left(K-1\right)}) 
\end{equation}

In a semi-supervised context,  the parameters $\left\{\Wvect^{(0)}, \Wvect^{(1)}, \ldots, \Wvect^{(K-1)} \right\}$ are learned by minimizing the Cross-Entropy loss function:

\begin{equation}\label{ls1}
{\mathcal{L}}_{CE}=-\sum_{i\in L}\sum_{j=1}^{c}{Y_{ij} \, ln   \, Z_{ij}}
\end{equation}
here  $\Yvect$ is the ground-truth label matrix and $L$ refers to the set of labeled samples.

\begin{figure*}[h!]
\centering
\includegraphics[width=1.0\textwidth]{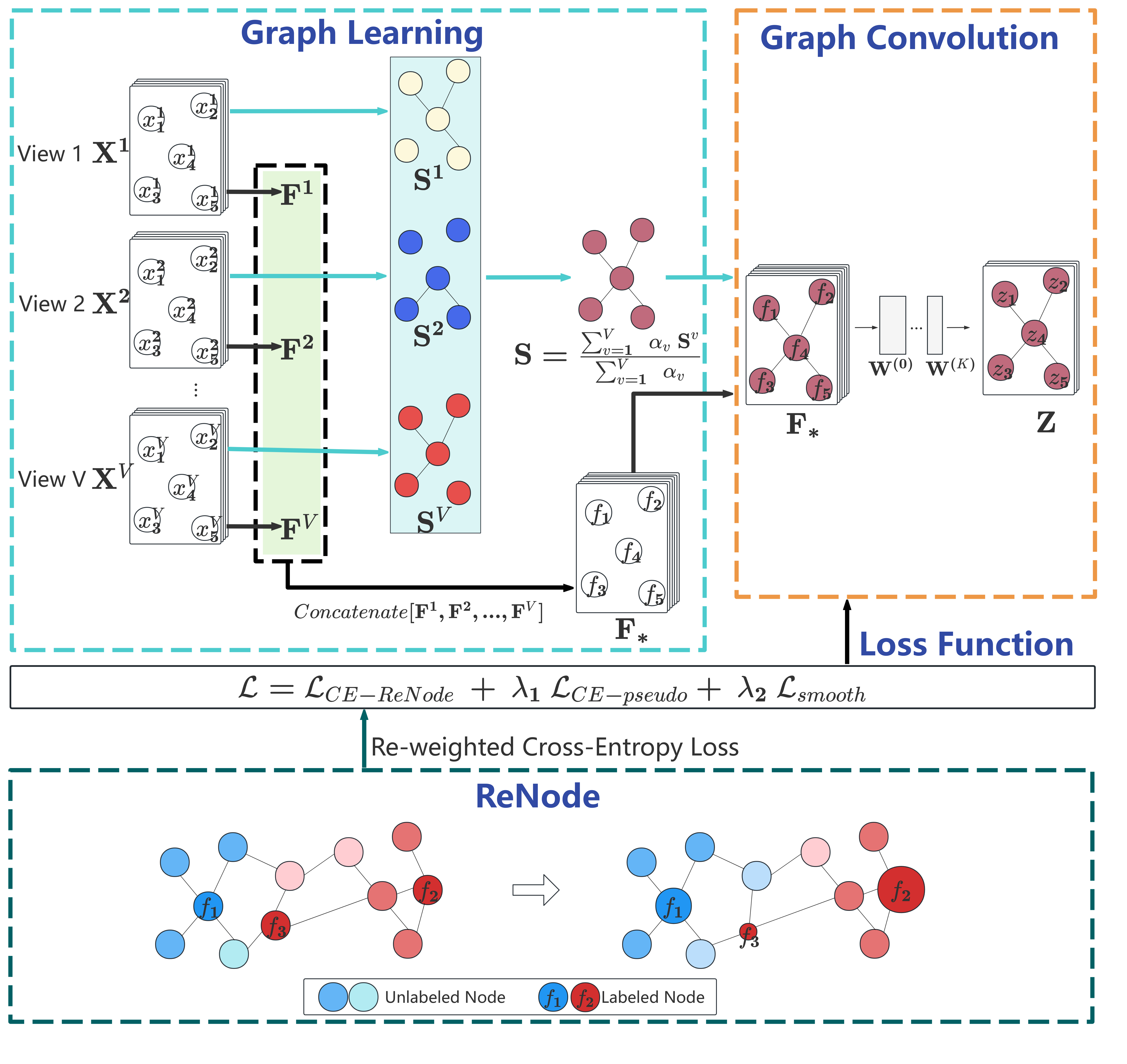}
\caption{The framework of our proposed RSGSLM. The model comprises three distinct modules: the first module is dedicated to graph learning and feature transformation, the second module handles topological imbalance adjustment, and the third module performs graph convolution.}
\label{figfuse}
\end{figure*}

%%%%%%%%%%%%%%%%%%%%%%%%%%%%
\section{Proposed approach}

Extending GCNs to handle multi-view data presents significant challenges, particularly when dealing with image data where inherent graph structures are not readily available. To tackle this issue, we introduce an innovative model named Re-node Self-taught Graph-based Semi-supervised Learning for Multi-view Data (RSGSLM). As illustrated in Figure \ref{figfuse}, our approach is structured into three interconnected modules:
(i) Joint Graph Learning and Feature Transformation (abbreviated as Graph Learning in Figure \ref{figfuse}): The first module focuses on simultaneously learning the graph structure and transforming features within each view.
(ii) Topological Imbalance Adjustment (abbreviated as ReNode in Figure \ref{figfuse}): The second module tackles the problem of topological imbalance by adjusting the weights of labeled samples according to their proximity to class boundaries, ensuring a more balanced representation in the learning process.
(iii) Unified GCN Architecture (abbreviated as Graph Convolution in Figure \ref{figfuse}): The third module employs a unified GCN architecture to generate soft labels. The combined loss function incorporates several elements, such as Re-weighted Cross-Entropy loss, Pseudo-Label loss, and Label Smoothness loss, which together enhance the model's performance.

By combining these modules, RSGSLM effectively addresses the challenges of multi-view data analysis and improves semi-supervised learning outcomes.  Each module is described in the following subsections.

\subsection{Graph learning}

The graph learning process is divided into two steps: first, a graph $\Svect^v$ is generated for each view, and then all $V$ graphs are adaptively merged into a single graph $\Svect$.

\subsubsection{Individual graph learning $\Svect^v$}

This section outlines the process of constructing a graph for each individual view.

The input matrices \{$\Xvect^v,v=1,2,\ldots, V $\} consist of $V$ views, where each view $\Xvect^v = [\xvect_1^v; \ \xvect_2^v; \ldots; \xvect_n^v] \in \mathbb{R}^{n \times d_v}$.
Initially, $\Xvect^v$ are normalized using the $\ell_2$ norm, ensuring that each column vector becomes a unit vector.

The objective is to generate $V$ unique graphs $\Svect^v$ that capture the pairwise relationships between the samples in each individual $\Xvect^v$.
Each graph matrix $\mathbf{S}^v$ is derived from its corresponding data $\mathbf{X}^v$ using a shallow and flexible semi-supervised approach that leverages labeled samples. Inspired by the MVCGL method \cite{2021MVCGL}, which proposed a shallow scheme to estimate both the graph and a versatile feature mapping, we have adapted this framework. In our method, we apply the framework separately to each view to derive the graph for that view.

Each graph $\mathbf{S}^v$ is obtained by minimizing this function:

\begin{equation}
\begin{split}
\label{Sv}
\min\limits_{\Svect^v,\Fvect^v,\Qvect^v,\bvect^v} 
 Trace( {\Xvect^v}^T    \Lvect^v  \Xvect^v) + 
\eta \,  Trace({\Fvect^v}^T \Lvect^v \Fvect^v) \\
+Trace((\Fvect^v-\Yvect)^T \Uvect(\Fvect^v-\Yvect)) 
+ \gamma \left\|\Svect^v\right\|^2 \\
+ \mu (\left\|\Qvect^v\right\|^2+
\alpha \left\| {\Xvect^v} \Qvect^v + {\bf 1} \, \bvect^{vT} - \Fvect^v \right\|^2)   \\
 \quad s.t.   \quad  0 \leq \Svect_{ij}^v \leq 1,   \quad   \sum_{j=1}^N S_{ij}^v=1
\end{split}
\end{equation}

\textcolor{black}{Where $\Fvect^v$ represents the soft label predictions of $\Xvect^v$, $\Lvect^v$ is the normalized Laplacian matrix derived from $\Svect^v$, $\Qvect^v$ represents the view-specific projection matrix,
the vector $\bvect^v$ denotes the bias item, the diagonal matrix $\Uvect$ assigns non-zero values to labeled samples while leaving the rest at zero.
$\eta$, $\gamma$, $\mu$, and $\alpha$ are balancing factors, $\mathbf{1}$ stands for a column vector with $n$ ones.
It's important to note the distinction between $\Fvect^v$ (from equation (\ref{Sv})) and $\mathbf{Z}$ (from equation (\ref{z1})). $\Zvect$ represents the output of GCN, whereas $\Fvect^v$ pertains to the shallow prediction of the labels for view $v$. 
This approach is especially effective for semi-supervised learning (SSL). The first two elements incorporate smoothing for both features and labels within the resulting graph structure, while the third element promotes label alignment. The final two elements are essential for effectively transforming features into the labeling domain, significantly enhancing the construction of robust semi-supervised methods. As a result, utilizing this method for graph construction is more suitable for semi-supervised models than for simple unsupervised learning techniques like KNN that  solely relies on pairwise similarities or distances.}

The optimization of equation~(\ref{Sv}) simultaneously constructs the view-specific graph $\Svect^v$ and the soft labels $\Fvect^v$. In our approach, $\Fvect^v$ serves as a compact linear transformation of $\Xvect^{v}$.

\subsubsection{Fused graph $\Svect$}

The fused graph \(\Svect\) is computed as an adaptive convex combination of the individual graphs, expressed as:

\begin{equation}
\label{Sfuse}
\Svect=\frac{\sum_{v=1}^{V}   {\alpha_v} \, \Svect^v}{\sum_{v=1}^{V} {\alpha_v}}
\end{equation}
where $\alpha_v$ represents the weight attributed to view $v$, and it is determined based on the corresponding smoothing item:

\begin{equation}
\label{alphav}
\alpha_v= \sqrt{Trace({\Xvect^v}^T {\Lvect}^v \Xvect^v)}
\end{equation}

According to equations~(\ref{Sfuse}) and (\ref{alphav}), views with lower data smoothness receive higher weights. As a result, these views have a greater influence on the unified graph and the final label smoothing term $\mathcal{L}_{smooth}$, as used in the GCN learning equation (\ref{gcnmr}).

\subsection{Re-weight nodes}

\textcolor{black}{Additionally, we modify the weights of the labeled nodes as described in \cite{chen2021topology}, a method initially designed for single-view problems, and extend its application to tackle multi-view challenges.}
In this study, the labeled nodes are assigned weights based on their proximity to the class boundaries, determined by computing the Totoro score. To assess the influence distribution of each labeled node, we compute the Personalized PageRank matrix  $\Pvect$ as follows:

\begin{equation} \label{pagerank}
\Pvect=\xi \, (\Ivect -(1-\xi)\Dvect^{-\frac{1}{2}}\Svect\Dvect^{-\frac{1}{2}} )^{-1} 
\end{equation}
here $\Ivect$ denotes the identity matrix and $ \xi \in (0,1]$ represents the probability of restarting the random walk. 

The Totoro value $T_i$ of the node $i$ quantifies its topological closeness to the centroid of its respective class and is determined by:

\begin{equation} \label{Ti}
\begin{aligned}
T_i= \mathbbm{E}_{x \sim \Pvect_i:}[\sum_{j\in [1,c],j \neq y_i}\frac{1}{\left |C_j\right|}\sum_{k\in C_j}{P_{k,x}}]
\end{aligned}
\end{equation}

In this equation, $y_i$ is the ground-truth label, $c$ presents the number of classes, and ($C_1, C_2, \ldots, C_c$) represents the training sets for each class. A smaller Totoro value indicates a higher relevance of labeled sample $i$. Consequently, the weight $ w_i$ is obtained by the cosine wave-based mapping:

 \begin{equation} \label{wv}
\begin{aligned}w_i&=\ w_{min}
&+\frac{1}{2}\ \left(w_{max}-w_{min}\right)\left(1+\cos{\left(\frac{Rank\left(T_i\right)}{\left|L\right|}\pi\right)}\right),
\ i\in  L
\end{aligned}
\end{equation}

In this formula, $w_{min}$ and $w_{max}$ are the lower and upper bound of the weight correction factor respectively, and rank($T_i$) represents the ranking order of $T_i$ within the labeled samples. 
By applying the re-weighted labeled samples, a novel cross-entropy loss function is derived according to equation~(\ref{ls1}). This re-weighted cross-entropy loss $\mathcal{L}_{CE-ReNode}$ is computed using:

\begin{equation}\label{CE-ReNode}
\mathcal{L}_{CE-ReNode}=-\frac{1}{\left|L\right|}\sum_{i\in L}w_i\sum_{j=1}^{c}{Y_{ij} \, ln Z_{ij}}
\end{equation}

\subsection{Graph convolution}

The input feature matrix $ \Fvect_\ast = [ \Fvect^1, \ \Fvect^2, \ldots, \Fvect^V]  \in \mathbb{R}^{n \times (c \times V)}$ is created by concatenating $\Fvect^v$ derived in each view.

Using $ \Fvect_\ast$ and the graph $\Svect$, the GCN model undergoes training. Propagation via hidden layers in each layer is:
\begin{equation}\label{h1} \Fvect^{(k+1)}=\sigma(\Dvect^{-\frac{1}{2}}\Svect\Dvect^{-\frac{1}{2}}\Fvect^{\left(k\right)} \Wvect ^{\left(k\right)})
\end{equation}
here $\Fvect^{(0)}=\Fvect_\ast$,  and the last layer is:
\begin{equation}\label{z1f+1}
\Zvect = softmax(\Dvect ^{-\frac{1}{2}} \Svect \Dvect ^{-\frac{1}{2}} \Fvect ^{\left(K-1\right)} \Wvect ^{\left(K-1\right)}) 
\end{equation} 

\subsection{Loss function}

In addition to the loss function \(\mathcal{L}_{CE-ReNode}\) described in Section 4.2, our proposed method also incorporates two additional loss terms: the pseudo-labels loss and the label smoothing loss. These will be explained in detail in this section.

\subsubsection{Pseudo-labels loss}

Equation (\ref{ls1}) calculates the loss for the labeled samples. To increase the amount of supervision during training, we can also use the predicted labels of unlabeled samples.

 By doing so, we can generate a corresponding loss for the unlabeled portion of the data. Specifically, we use equation (\ref{CEp}) to compute the loss for the unlabeled samples, utilizing Pseudo-Labels:

\begin{equation}\label{CEp}
{\mathcal{L}}_{CE-pseudo}=- w_p  \sum_{i\in L_u}  \,   \sum_{j=1}^{c}{Y^p_{ij} \, ln   \, Z_{ij}} 
\end{equation}
where $L_u$ represents the set of unlabeled samples, and $u = |L_u|$ denotes the number of unlabeled samples. The matrix of pseudo labels $\Yvect^p   \in \mathbb{R}^{u  \times c }$  is composed of the output predictions, $\Zvect$, from the deep network obtained during the previous epoch.

 It is important to note that the pseudo label matrix $\Yvect^p   \in \mathbb{R}^{u  \times c }$ differs from the ground-truth label $\Yvect \in \mathbb{B}^{n  \times c }$. In $\Yvect \in \mathbb{B}^{n  \times c }$ the elements are binary (0 or 1), whereas in $\Yvect^p$, the elements are real values between 0 and 1, representing the probability that a sample belongs to a particular class. These probabilities can be interpreted as confidence levels. Therefore, when $Y^p_{ij}$ has a high value (indicating high confidence in the prediction), the term $Y^p_{ij} \, ln   \, Z_{ij}$ is also large, and the opposite is true when $Y^p_{ij} \, ln \, Z_{ij} $ is small.

The value of $w_p$ increases as training progresses. At the beginning of training, the pseudo labels generated by the deep model may be inaccurate, so the weight of the pseudo-label loss ${\mathcal{L}}_{CE-pseudo}$, $w_p$ should be relatively small. However, as training advances and the model becomes more capable of producing accurate pseudo labels, the value of $w_p$ should increase accordingly.
This weight can follow a linear temporal schedule, as described below:
\begin{equation}\label{wp}
w_p=\frac{epoch - 1}{Max}
\end{equation}
here $Max$ represents the maximum number of epochs in the training process. In our experiments, we set $Max=2000$. This means that when $epoch=1$, $w_p=\frac{epoch - 1}{Max}=(1-1)/2000=0$, and when $epoch=2$, $w_p=(2-1)/2000=0.0005$. At epoch=1001, $w_p=1000/2000=0.5$, and finally, at epoch=2000, $w_p=1999/2000=0.9995$, which is close to 1.

In summary, ${\mathcal{L}}_{CE-pseudo}$ uses pseudo labels to enhance the cross-entropy loss by taking into account the confidence of the predicted labels and the current model.
Since the values in $\Yvect^p$ reflect confidence levels, the weight of each prediction is dynamically adjusted based on its confidence, removing the need to set a fixed threshold for pseudo labels. This ensures that no pseudo-label information is discarded due to low confidence.
Additionally, the flexible weighting factor $w_p$ is introduced to address the issue of inaccurate pseudo labels during the early stages of deep learning.

\subsubsection{Label smoothness loss}

The manifold regularization loss  $\mathcal{L}_{smooth}$ \cite{kejani2020graph}  is defined in equation~(\ref{gcnmr}), and it enhances the smoothness of predicted labels among all samples by employing $\Svect$.

\begin{equation}\label{gcnmr}
\mathcal{L}_{smooth}=  \frac{1}{2} \sum_{i=1}^n \sum_{j=1}^n  ||\Zvect_i - \Zvect_j   ||^2 S_{ij} = Trace(\Zvect^T\Lvect \Zvect)
\end{equation}

\subsubsection{Loss of RSGSLM}

The loss function of RSGSLM includes three components: Re-weighted Cross-Entropy loss $\mathcal{L}_{CE-ReNode}$, Manifold Regularization loss $\mathcal{L}_{smooth}$, and the Pseudo-Labels loss $\mathcal{L}_{CE-pseudo}$.
The Re-weighted Cross-Entropy loss $\mathcal{L}_{CE-ReNode}$ is obtained from equation~(\ref{CE-ReNode}).

The loss function of RSGSLM is:
\begin{equation}
\begin{split}
\label{second}
\mathcal{L}_{RSGSLM}=\underbrace{\mathcal{L}_{CE-ReNode}}_{Semi-supervised \ item} 
+ \underbrace{\lambda_1 \, \mathcal{L}_{CE-pseudo}}_{Pseudo- Label} 
+ \underbrace{\lambda_2 \, \mathcal{L}_{smooth}}_{unsupervised \ item}
\end{split}
\end{equation}
where $\lambda_1$ and $\lambda_2$ are balancing parameters between $\mathcal{L}_{CE-ReNode}$, $\mathcal{L}_{CE-pseudo}$ and $\mathcal{L}_{smooth}$.

The key steps of our RSGSLM model are shown in Algorithm \ref{alg:algorithm1}. The code can be found at \footnote{Source code: https://github.com/BiJingjun/RSGSLM.}.

\begin{algorithm}[t]
	\caption{RSGSLM.}
	\label{alg:algorithm1}
	\KwIn{Data \{$\Xvect^v,v=1,2,\ldots, V $\} from multiple perspectives, along with a subset of $l$ labeled samples}

	\KwOut{The fused graph $\Svect$, the soft label prediction $\Zvect$}  
	\BlankLine
 \begin{enumerate}
 
\item Apply the $\ell_2$ normalization to each $\Xvect^v$ to convert its column vectors into unit vectors.

\item For each view $v = 1, \ldots, V$, compute individual graphs $\Svect^v$ and their respective projected features $\Fvect^v$ according to equation~(\ref{Sv}).

\item Compute the fused graph $\Svect$ using  equation~(\ref{Sfuse}).

\item Concatenate the projected features $\Fvect^v$ from each view $\Fvect_\ast = [\Fvect^1, \Fvect^2, \ldots, \Fvect^V] \in \mathbb{R}^{N \times (V \times c)}$.

\item Calculate the weight of each labeled node by equation~(\ref{wv}).

\item Input the fused graph $\Svect$ and the concatenated $\Fvect_\ast$ into a 2-layer GCN.

\item Train this GCN by minimizing the global loss $\mathcal{L}_{RSGSLM}$ according to  equation~(\ref{second}) with a gradient descent technique. At each epoch, obtain the pseudo-labels $\Yvect^p$ from the output $\Zvect$ generated at the previous epoch and compute ${\mathcal{L}}_{CE-pseudo}$ using $\Yvect^p$ and $w_p$ according to equations~(\ref{CEp}) and ~(\ref{wp}).

\item After completing the training, the soft labels $\Zvect$ are obtained.
\end{enumerate}
\end{algorithm}

\section{Experimental procedure}

In this section, we provide a thorough introduction to the experimental procedure. We will cover an overview of the competition models used, the experimental datasets implemented, as well as the parameter and training settings applied.

\subsection{Competing models}

In this study, we compare the performance of RSGSLM with ten multi-view semi-supervised models. Two of these models (GCN-$\Xvect_\ast$ and GCN-multi) are baseline methods, and the other eight methods are advanced approaches in SSL with multiple views, including deep methods and shallow methods, which were presented in detail in our related work.

The two baseline models based on GCN are used to verify whether our proposed techniques achieve higher accuracy.

\textbf{GCN-$\Xvect_\ast$} generates the unified graph $\Svect$ by solving problem~(\ref{Sv}) with the concatenated $\Xvect_\ast = [ \Xvect^1, \ \Xvect^2, \ldots, \Xvect^V] \in \mathbb{R}^{N \times (d_1+d_2+\ldots+d_v)}$.
It then applies a standard GCN \cite{kipf2016semi} using $\Xvect_\ast$ and the constructed graph $\Svect$.

\textbf{GCN-multi} executes GCN \cite{kipf2016semi} individually for each view. Each run assigns $\Xvect^v$ as input features, with the corresponding $\Svect^v$ constructed by function~(\ref{Sv}), and obtains output label matrices $\Zvect^v$. To obtain the final prediction, we calculate the mean of these $V$ outputs $\Zvect = \frac{1}{V} \sum_{v=1}^V \Zvect^v$.

\subsection{Datasets}

\begin{table}[]
\caption{Dataset details.}
\label{datasets8}
\centering
\resizebox{0.9\linewidth}{!}{
\begin{tabular}{lllll}
\hline
View                        & ORL                                        & Scene                                                             & ALOI                      \\ \hline
1                           & 512-D GIST                                 & 512-D GIST                              & 64-D RGB  \\
2                           & 59-D Local Binary Pattern                  & 432-D Color moment          & 64-D HSV \\
3                           & 864-D HOG                                  & 256-D HOG                                          & 77-D Color similarities   \\
4                           & 254-D Centrist                             & 48-D Local Binary Pattern                            & 13-D Haralick features    \\
5                           & --                                         & --                                          & --                        \\
6                           & --                                         & --                                                   & --                        \\
\# Samples                  & 400                                        & 2688                                                                    & 1079                      \\
\# Classes                  & 40                                         & 8                                                                       & 10                        \\
\# Train of each class      & 3                                          & 15                                                                        & 20                        \\
\# Validation of each class & 2                                          & 15                                                                         & 20                        \\
Data type                  & Face image                                 & Scene image                                                         & Object image              \\ \hline
                            &                                            &                            &                                                     &                           \\ \hline
View   & Youtube                            & Handwritten                  &MNIST\\ \hline
1     & 2000-D Cuboids histogram     & 240-D Pixel averages                &2048-D Resnet50 \\
2     & 1024-D Hist motion estimate   & 76-D Fourier coefficients          &4096-D VGG16 FC1\\
3     & 64-D HOG features                   & 216-D Profile correlations    & -- \\
4     & 512-D MFCC features              & 47-D Zernike moments             & --\\
5     & 64-D Volume streams               & 64-D Karhunen-Love coefficients  & --\\
6     & 647-D Spectrogram streams       & 6-D Morphological features          & --\\
\# Samples & 2000                       & 2000                                 & 10000\\
\# Classes & 10                             & 10                               &10\\
\# Train of each class  & 20          & 5                                      &20\\
\# Validation of each class & 20        & 3                                    &20\\
Data type  & Video data                  & Digit image                         &Digit image\\ \hline
\end{tabular}
}
\end{table}

We assess the efficacy of the RSGSLM model across five benchmark datasets containing multi-view image data.
These are ORL (facial image) \footnote{http://cam-orl.co.uk/facedatabase.html}, Scene (outdoor scenes) \footnote{https://scholar.googleusercontent.com/scholar?q=cache:Dxo2Hbfln2sJ: scholar.google.com/hl=enas-sdt=0,5}, 
ALOI (object image) \footnote{https://elki-project.github.io/datasets/multi\_view}, 
Youtube (video) \footnote{http://archive.ics.uci.edu/ml/datasets}, Handwritten (handwritten digits) \footnote{https://archive.ics.uci.edu/ml/datasets/Multiple+Features}, and MNIST (digit image) \footnote{http://yann.lecun.com/exdb/mnist/}.
Further information regarding these datasets is available in Table \ref{datasets8}.

\subsection{Parameter setting}

We obtained the code for MVCGL, AMSSL and LACK from their authors, while the code for JCD\footnote{https://github.com/ChenpingHou/Joint-Consensus-and-Diversity-for-Multi-view-Semi-supervised-Classification}, DSRL\footnote{https://github.com/chenzl23/DSRL}, LGCN-FF \footnote{https://github.com/chenzl23/LGCNFF}, IMvGCN\footnote{https://github.com/ZhihaoWu99/IMvGCN}, and TUNED \footnote{https://github.com/JethroJames/TUNED} was downloaded from GitHub. Each of these models was evaluated using the default parameters. For a fair comparison, the parameters of the two baseline methods were set identically to those used in the RSGSLM model.

When generating the graph (as defined in equation (\ref{Sv})), the parameters were set as $\alpha=0.003$, $\gamma=0.003$, $p=2$, $\mu=100$ and $\eta=5$. $\xi$ is 0.15 in equations (\ref{pagerank}), $w_{min}$ is 0.5 in equations (\ref{wv}), and ($w_{max}-w_{min}$) were varied in the range \{0.1, 0.2, 0.3, 0.4, 0.5, 0.6, 0.7, 0.8, 0.9, 1.0, 2.0, 3.0, 4.0, 5.0, 10.0\}.

A two-layer GCN architecture was used for the GCN methods, with the training process capped at a maximum of 2000 epochs and an early-stop criterion activated after 100 epochs. The dimensionality of the hidden layer in GCN was chosen from the set \{12, 14, 16, 18, 20, 22, 24, 26, 28, 30, 32, 34, 36, 38\}, while the initial learning rates were chosen from \{0.005, 0.01, 0.05, 0.1\}.
$\lambda_1$ and $\lambda_2$ in equation (\ref{second}) were varied in the range \{1e-9, 1e-8, 1e-7, 1e-6, 1e-5, 1e-4, 1e-3, 1e-2, 1e-1, 1, 10, 100, 1000\}.

\subsection{Training settings}

Table \ref{datasets8} contains the number of training or validation samples, the rest is reserved for testing. Each model was run ten times per dataset. The dataset was randomly split for each run into training and testing sets. All models employed identical same ten training/testing splits to guarantee a consistent and fair comparison.

\section{Results}

\begin{table}[]
\caption{Accuracy of the ten models on five multi-view image datasets.}
\label{10result}
\centering
\resizebox{0.9\linewidth}{!}{
\begin{tabular}{llllllll}
\hline
\textbf{Method}      & ORL           & Scene       & ALOI         &Youtube       & Handwritten  & MNIST\\ \hline
GCN-$\Xvect\ast$     & 93.14±2.3     & 74.57±1.4   & 98.20±0.8    & 41.59±2.0    & 92.31±1.6    & 93.39±0.6\\
GCN-multi            & 92.70±2.1     & 64.43±3.1   & 96.70±0.4    & 47.28±1.3    & 91.47±1.4  & 92.82±0.6\\
MVCGL \cite{2021MVCGL}&  94.67±2.3    & 76.28±1.0   & 94.49±0.5    &42.57±1.1    & 95.38±1.3  & 95.53±0.4\\
AMSSL \cite{AMSSL2021}& 93.10±2.2     & 54.22±2.3   & 49.30±3.5    & 27.03±1.4   & 89.42±2.3  & 93.27±0.2  \\
JCD \cite{JCD}        & 93.65±2.5     & 45.16±3.7   & 98.01±0.3    & 11.46±2.8   & 95.95±0.7  & \textbf{96.42±0.3}\\
DSRL \cite{DSRLwang}  & 92.50±1.8     & 68.08±13.0  & 94.00±1.1    & 43.29±1.3   & 90.74±1.6  & 91.39±0.4\\ 
LGCN-FF \cite{LGCNFF2023}& 64.90±9.8  & 61.47±9.9   & 97.03±0.6    & 26.99±9.8   & 86.19±4.0  & 60.75±4.7\\
LACK \cite{lack2022semi} & 89.10±2.1  & 69.49±0.5   & 77.63±2.9    &  28.22±1.5  & 91.24±0.6  & 71.84±0.7\\
TUNED \cite{TUNED2025} & 96.75±2.5    & 79.05±0.9   & 97.91±1.0    &  52.32±3.0  & 96.42±0.7  & 40.07±1.3\\
IMvGCN \cite{ImvGCN2023} & 80.90±0.6  & 77.60±0.6   & 81.28±3.7    & 55.65±0.8   & 94.63±0.3  & 90.79±0.1\\ \hline
RSGSLM          & \textbf{97.30±1.2} & \textbf{79.91±0.3}  & \textbf{98.90±0.4}   & \textbf{56.00±1.3}   &\textbf{97.73±0.4} & 93.73±0.6\\ \hline

\end{tabular}
}
\end{table}

Table \ref{10result} presents accuracy results (average\% and standard deviation\%) of ten multi-view SSL models for six benchmark datasets. For clarity, the top values are highlighted in bold. RSGSLM shows exceptional performance, achieving the best accuracy on five datasets with consistent results across various datasets and splits, as evidenced by the low standard deviations. Whether it is simpler datasets such as ORL, ALOI, or Handwritten or more complex ones such as Scene or Youtube, RSGSLM consistently performs better.

\begin{table}[]
\caption{Optimal parameters.}
\label{parameter}
\centering
\resizebox{0.9\linewidth}{!}{
\begin{tabular}{lrrrrrr}
\hline
Dataset      & GCN hidden layer dimension & Learning rate & $\lambda_1$& $\lambda_2$& $w_{max}-w_{min}$ \\ \hline
ALOI           & 28     & 0.05      & 100     & 0.001   &3.0    \\
% NUS-WIDE    & 32           & 0.0001        & 0.05          & 100       \\
% MSRC-v1     & 120                 & 0.001         & 50     &200  &1  &0.001   \\
ORL            & 30     & 0.01      & 100 & 1    & 1.0 \\
Scene          & 28     & 0.005      & 1000    & 1  & 0.4   \\
Handwritten    & 34     & 0.001     & 1e-9  & 1  & 0.5   \\
Youtube        & 18     & 0.01      & 1000     & 10   & 0.9  \\ 
MNIST          & 20     & 0.01      & 1e-9 & 1000  & 0.1 \\ 
\hline
\end{tabular}
}
\end{table}

\begin{figure*}
  \centering
  \subfigure[Scene]{\includegraphics[width=\textwidth]{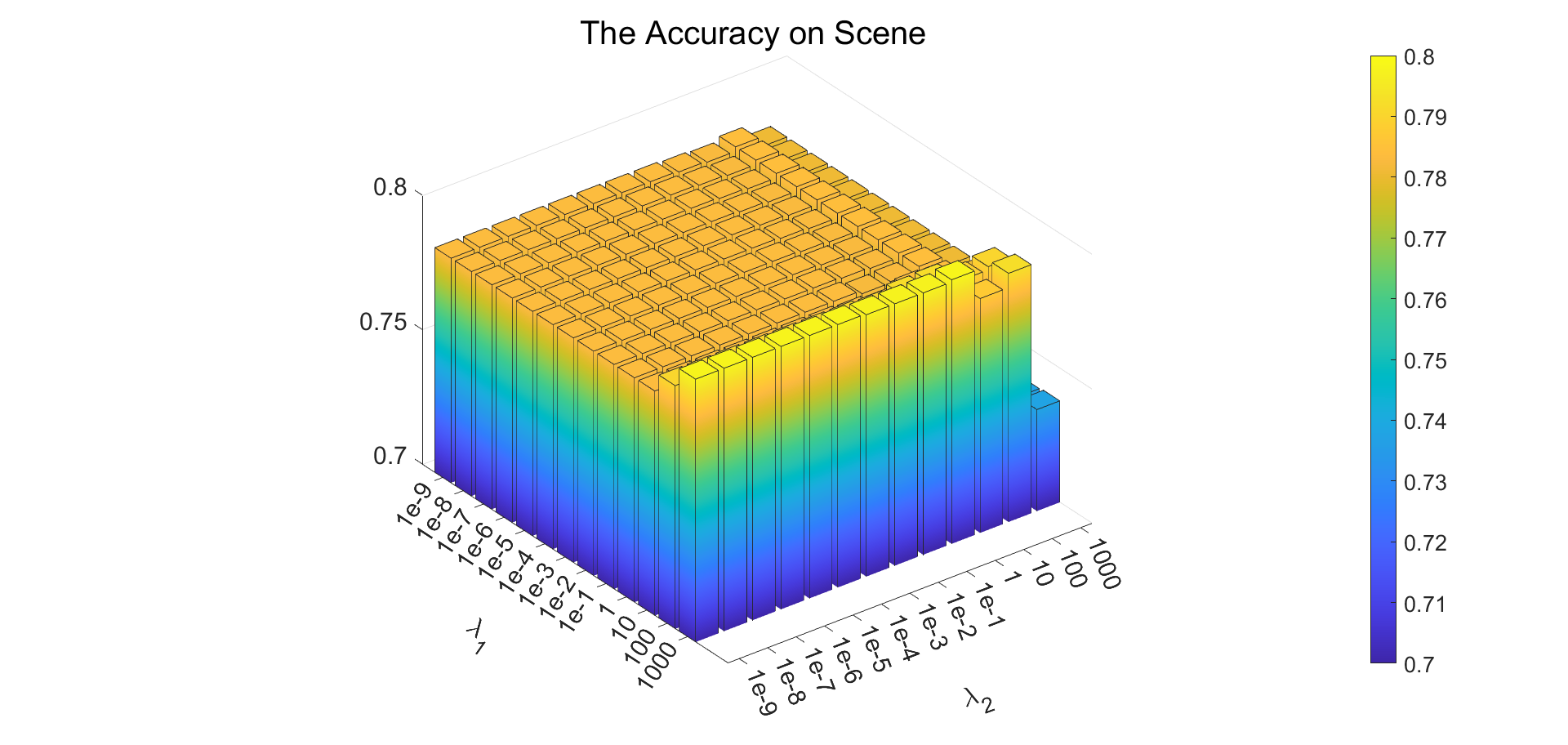}} 
  \subfigure[Youtube]  {\includegraphics[width=\textwidth]{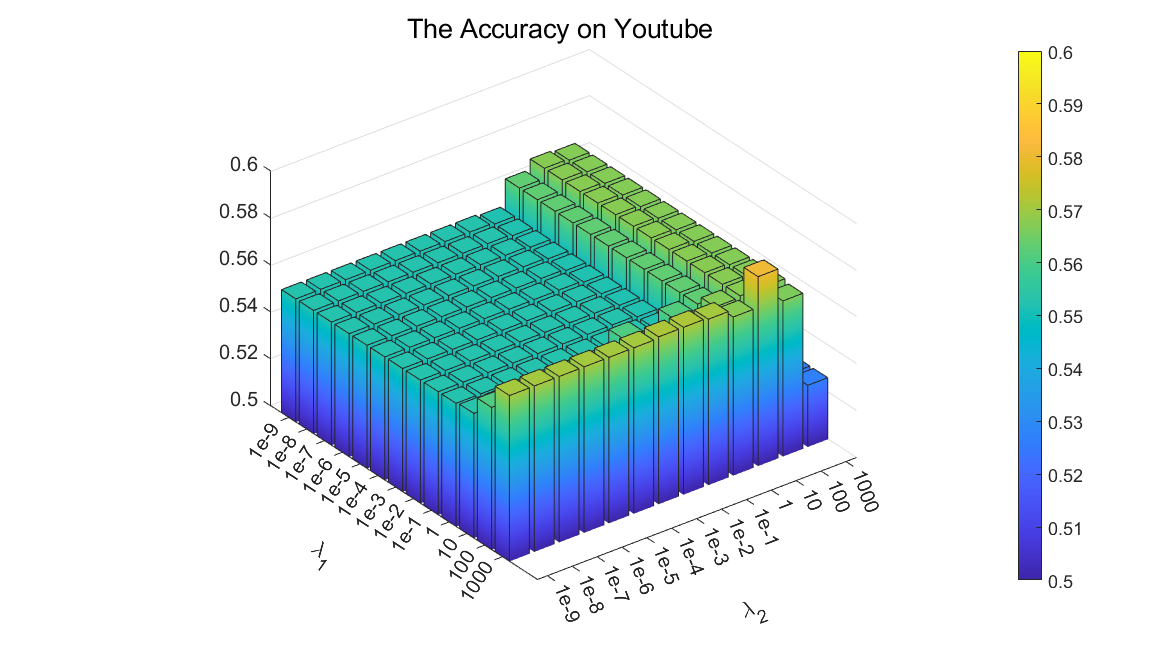}} 
  \caption{\textcolor{black}{Sensitivity to parameters $\lambda_1$ and $\lambda_2$ on the Scene and Youtube datasets.}}
  \label{parala}
\end{figure*}

\begin{figure*}
  \centering
  \includegraphics[width=0.9\textwidth]{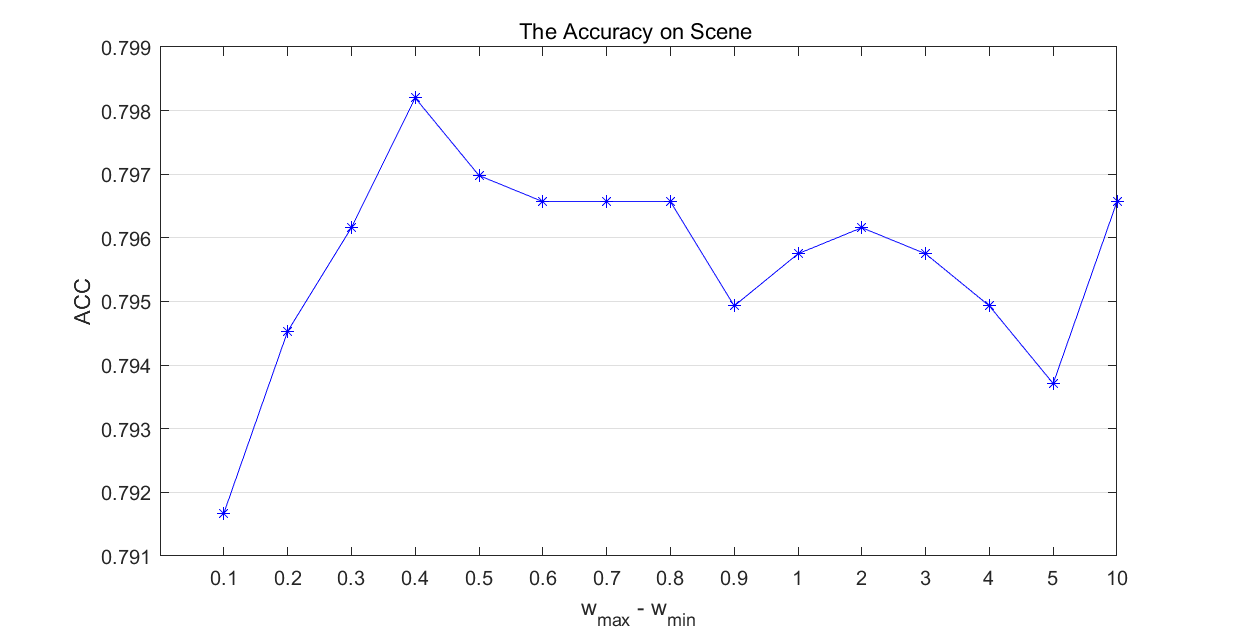}
  \caption{Sensitivity to parameter $w_{max}-w_{min}$ on the Scene dataset.}
  \label{paraw}
\end{figure*}

\section{Discussions}

This section offers a detailed examination of the performance of each individual component of our model using various analytical methods. The comprehensive analysis is organized into several key subsections: (i) Section 7.1 will present a sensitivity analysis of the model parameters; (ii) Section 7.2 will focus on a qualitative evaluation of the semi-supervised data embedding; (iii) Section 7.3 will feature two ablation studies; and (iv) Section 7.4 will analyze the computational complexity.

\subsection{Sensitivity to parameters}

We conducted parameter tuning to determine the optimal values for $\lambda_1$, $\lambda_2$, and $w_{max}-w_{min}$ in our RSGSLM model to maximize accuracy. Table \ref{parameter} summarizes the best parameter values selected for each dataset in our approach. These values represent the optimal parameters that achieved the highest accuracy in our experimental settings. However, this does not imply that other parameter values will necessarily result in poor accuracy. It is important to note that the optimal parameters for achieving the best accuracy may vary across different datasets.

\textcolor{black}{Moreover, changes in one parameter can affect the optimal values of other parameters. For instance, consider the challenging datasets Scene and Youtube. Figure \ref{parala}.(a) illustrates the relationship between the accuracy (for a single split) and the parameters $\lambda_1$ and $\lambda_2$ within the RSGSLM model on the dataset Scene when $w_{max}-w_{min}$ is fixed. Here, we use the optimum value $w_{max}-w_{min} = 0.4$ from Table 4 as an example. 
The figure shows that the effect on accuracy is relatively minor when $\lambda_1$ is less than 10, but the accuracy increases significantly when $\lambda_1$ exceeds 10. 
Since $\lambda_1$ corresponds to the loss function ${\mathcal{L}}_{CE-pseudo}$, this suggests that a larger contribution from ${\mathcal{L}}_{CE-pseudo}$ can be beneficial.
Conversely, as $\lambda_2$ increases, accuracy first rises, peaking at $\lambda_2 = 1$, and then declines sharply when $\lambda_2$ exceeds 100.
Because \(\lambda_2\) corresponds to the loss function \(\mathcal{L}_{smooth}\), this indicates that excessive penalization through \(\mathcal{L}_{smooth}\) can be counterproductive. Figure \ref{parala}.(b) illustrates a comparable scenario with the Youtube dataset.} 

Figure \ref{paraw} illustrates the relationship between accuracy (for a single split) and the parameters $w_{max}-w_{min}$ within the RSGSLM model, with fixed values of $\lambda_1$ and $\lambda_2$. Here, we use the optimum values $\lambda_1=1000$ and $\lambda_2=1$ from Table \ref{parameter} as an example. The figure shows that selecting an appropriate value for $w_{max}-w_{min}$ also significantly impacts model accuracy, with the highest accuracy achieved when $w_{max}-w_{min}=0.4$.

\subsection{Qualitative evaluation of the semi-supervised data embedding}

 \begin{figure}
  \centering
  \subfigure[$\Xvect_\ast$ on ORL]{\includegraphics[height=1.6in]{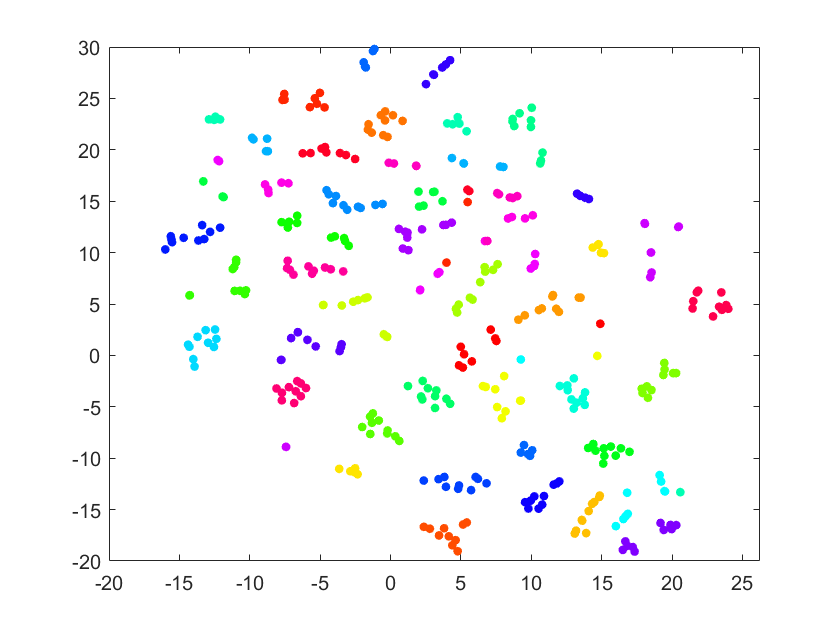}}
  \subfigure[$\Zvect$ on ORL]{\includegraphics[height=1.6in]{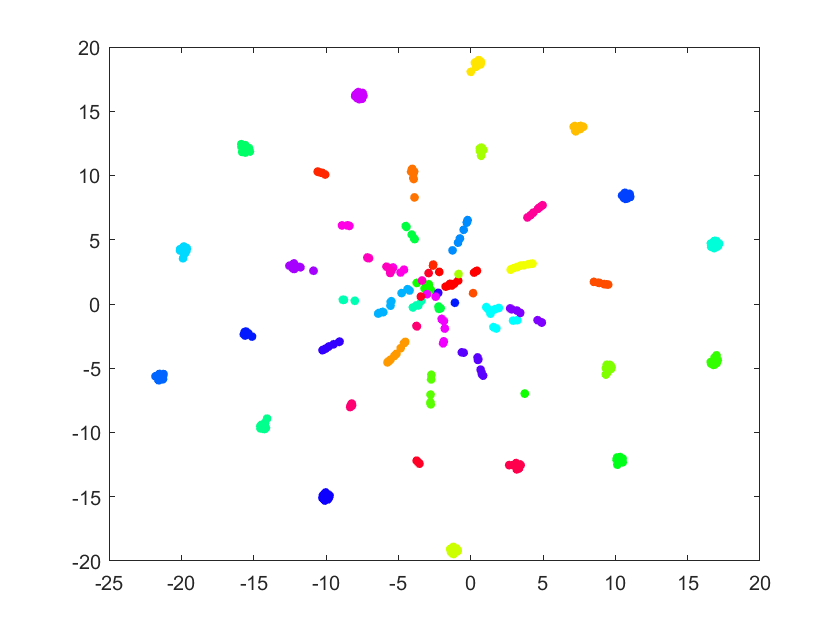}}  
  \subfigure[$\Xvect_\ast$ on Scene]{\includegraphics[height=1.6in]{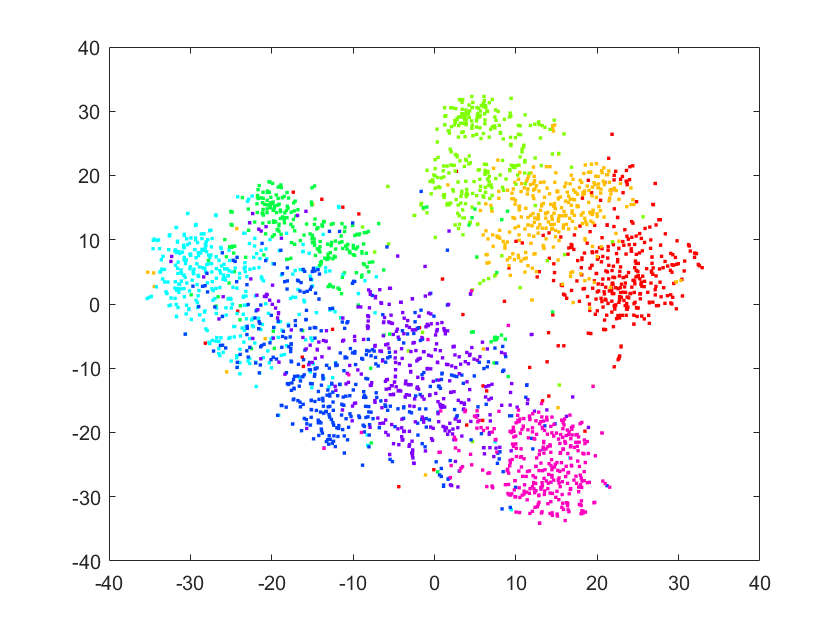}}
  \subfigure[$\Zvect$ on Scene]{\includegraphics[height=1.6in]{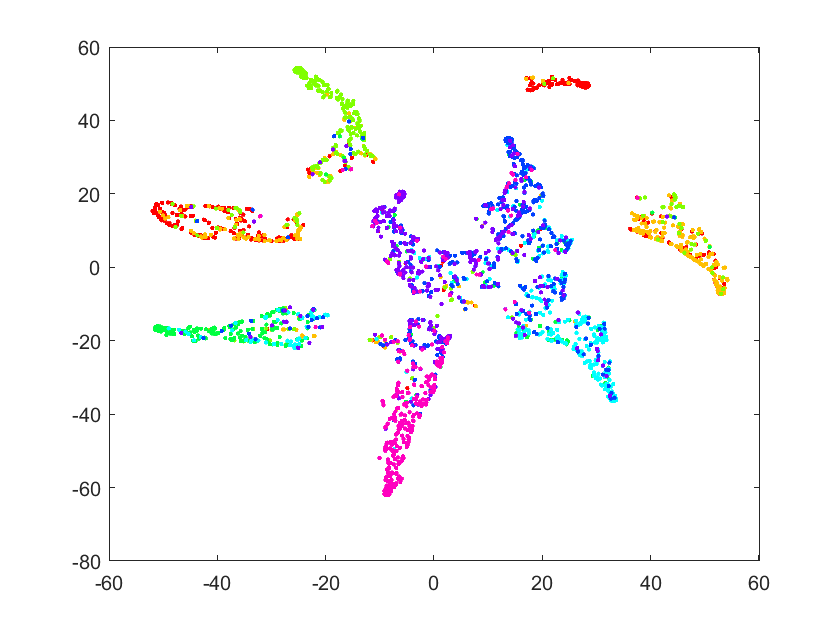}}
  \subfigure[$\Xvect_\ast$ on Youtube]{\includegraphics[height=1.6in]{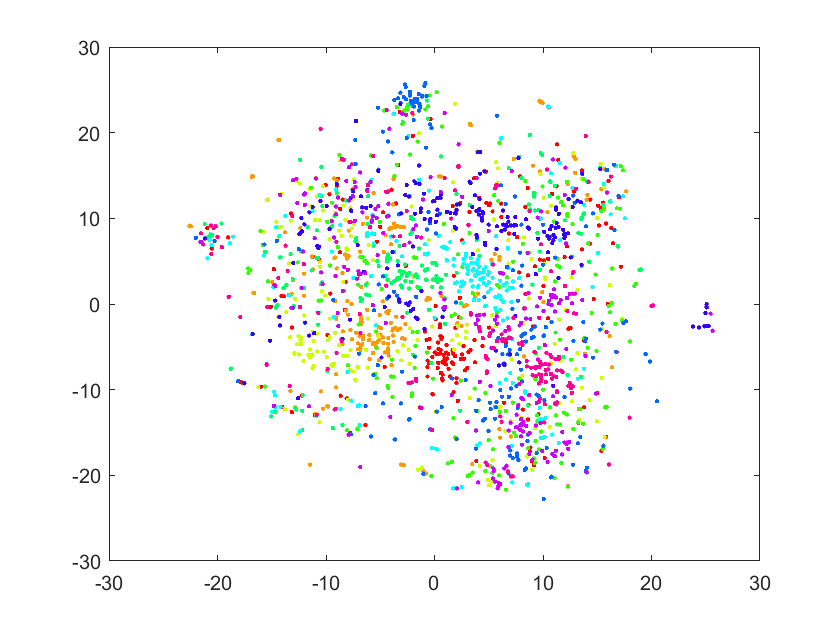}}
  \subfigure[$\Zvect$ on Youtube]{\includegraphics[height=1.6in]{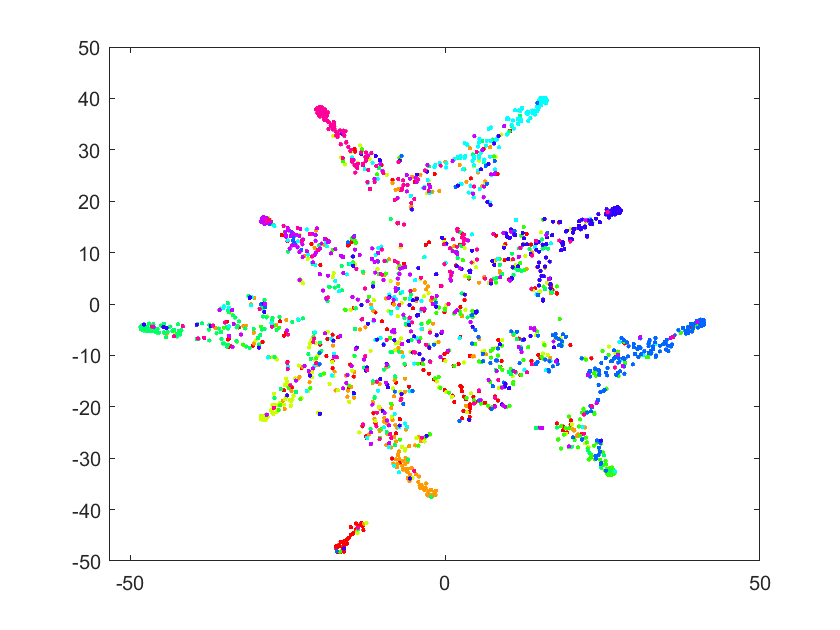}}
  \subfigure[$\Xvect_\ast$ on Handwritten ]{\includegraphics[height=1.6in]{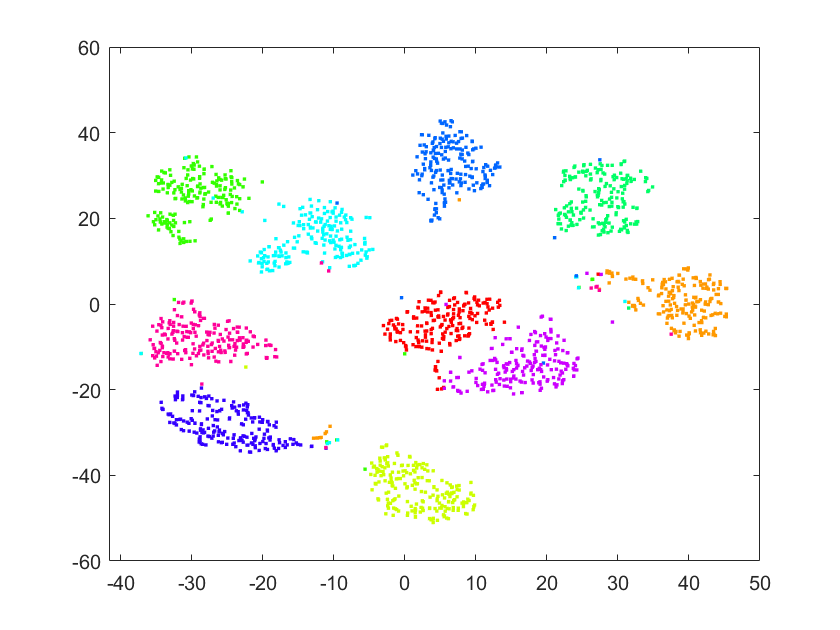}}
  \subfigure[$\Zvect$ on Handwritten ]{\includegraphics[height=1.6in]{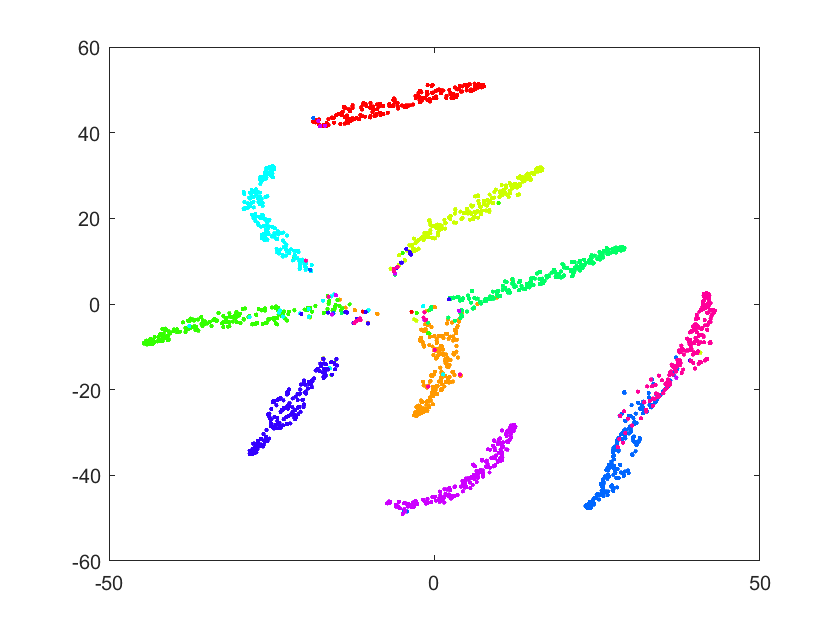}}     
  \caption{ The t-SNE visualization of $\Xvect_\ast$ and $\Zvect$ in RSGSLM on four datasets.}
  \label{FandZ}
\end{figure}

\begin{figure}
  \centering
  \subfigure[$\Xvect_\ast$ ]{\includegraphics[height=1.6in]{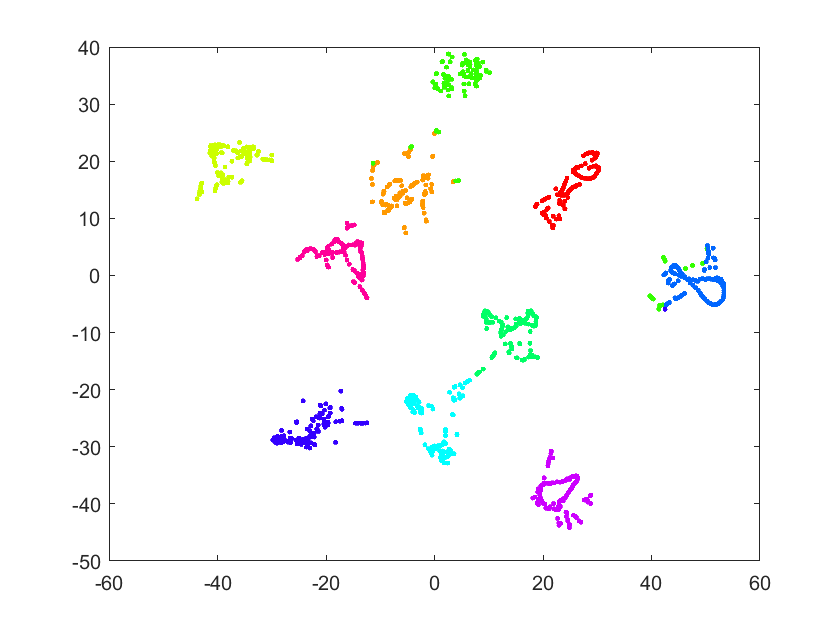}}
  \subfigure[Output of IMvGCN  ]{\includegraphics[height=1.6in]{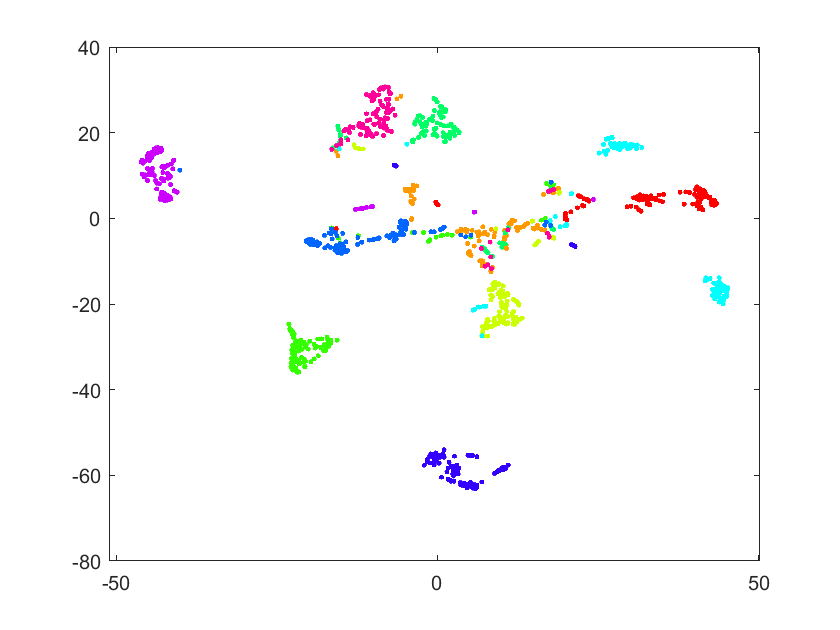}}  
  \subfigure[Output of LGCN-FF ]{\includegraphics[height=1.6in]{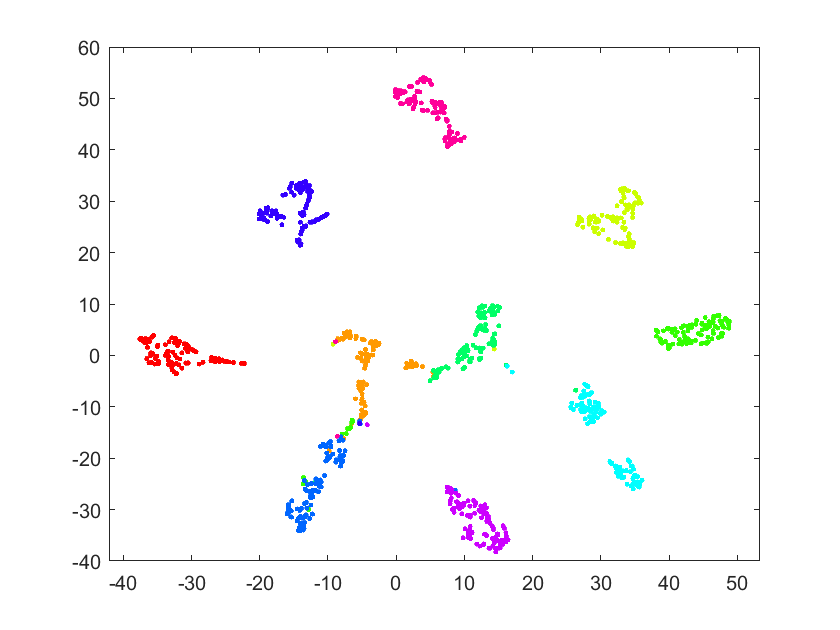}}  
  \subfigure[Output of our RSGSLM ]{\includegraphics[height=1.6in]{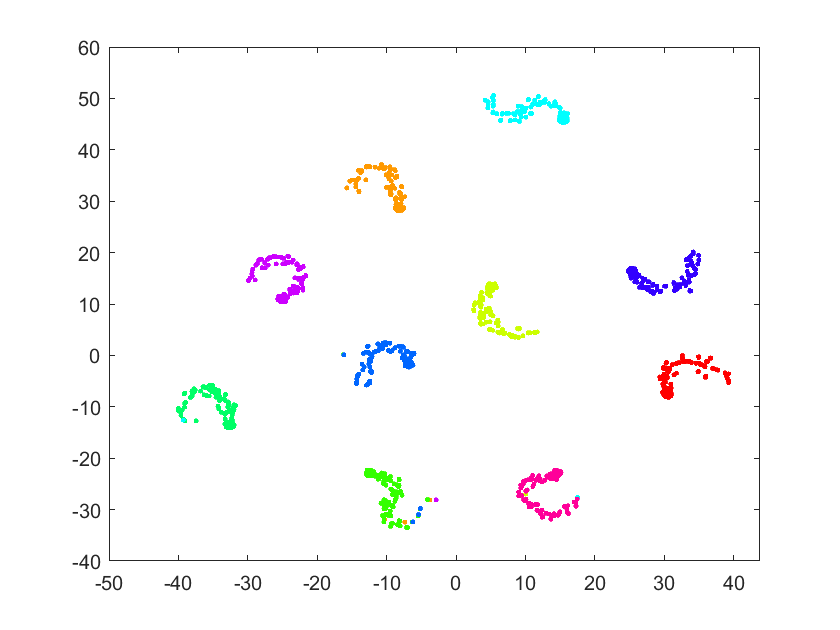}}  
  \caption{ The t-SNE visualization of $\Xvect_\ast$ and output of three models on ALOI.}
  \label{threeZ}
\end{figure}

To qualitatively evaluate the semi-supervised data embedding, we visualize both the original features and the output $\Zvect$ from the GCN. It begin by concatenating and normalizing $\Xvect^v$ from each view to create the concatenated $\Xvect_\ast = [ \Xvect^1, \ \Xvect^2, \ldots, \Xvect^V] \in \mathbb{R}^{N \times (d_1+d_2+\ldots+d_v)}$. True class labels are represented by different colors, and we apply t-SNE projection for visualizations,  with each dot representing one sample.
Figure \ref{FandZ} presents the t-SNE visualizations of $\Xvect_\ast$ and the learned representations $\Zvect$ on four datasets.

In the ORL dataset, the distribution of points in Figure \ref{FandZ}.(b) becomes smoother when using the embedding, with different classes being more effectively distinguished compared to Figure \ref{FandZ}.(a). A similar pattern is observed in the remaining datasets.

Figure \ref{threeZ} presents a comparative examination of the sample representations generated by RSGSLM and two competing GCN-based models on the ALOI dataset.
Figure \ref{threeZ}.(a) shows the original features $\mathbf{X}_\ast$.
Figure \ref{threeZ}.(d) displays the output $\Zvect$ generated by RSGSLM.
Figure \ref{threeZ}.(b) and Figure \ref{threeZ}.(c) present the outputs of the IMvGCN and LGCN-FF models, respectively.

It is clear that the distribution of the output from RSGSLM appears a smoother distribution in comparison to the output from the other two methods. Additionally, RSGSLM more clearly distinguishes almost all classes. In contrast, in Figure \ref{threeZ}.(b), the separation between the blue, orange, and red classes is less distinct. Likewise, in Figure \ref{threeZ}.(c), the separation between the blue, orange, and green data dots is indistinct and blurred. This indicates that RSGSLM provides better class separation and more coherent data embedding.

\subsection{Ablation study}

\textcolor{black}{In this section, we carried out three ablation experiments to evaluate the influence of the three loss terms, the scheduling strategies for $w_p$, and to examine the effects of substituting pseudo labels with ground-truth labels.}

\subsubsection{Three loss items}

\begin{table}[]
\caption{Ablation study of three loss items on the Scene dataset.}
\label{ablation}
\centering
\begin{tabular}{ccccc}
\hline
    &     $\mathcal{L}_{smooth}$ & $\mathcal{L}_{CE-pseudo}$& $w_i$ in $\mathcal{L}_{CE-ReNode}$  & Accuracy \% \\\hline
     
  1&    \ding{56}          & \ding{56}          & \ding{56}        & 78.59      \\
  2&   \ding{56}          &   \ding{56}        & \checkmark        & 78.79      \\     
  3&     \ding{56}        &     \checkmark     & \ding{56}         & 79.20    \\
  4&     \checkmark       &    \ding{56}       & \ding{56}         & 78.71      \\
  5&   \ding{56}          &   \checkmark       & \checkmark        & 79.77     \\ 
  6&     \checkmark       &    \ding{56}       & \checkmark        & 79.24     \\
  7&   \checkmark         &     \checkmark     & \ding{56}         &  79.21    \\
      
  8&     \checkmark       &     \checkmark     & \checkmark          & {\bf 79.82 } \\
\hline
\end{tabular}
\end{table}

We carried out an ablation study to investigate the effect of including the unsupervised term $\mathcal{L}_{smooth}$, the Pseudo-Labels loss $\mathcal{L}_{CE-pseudo}$, and the weight of the labeled node  $w_i$ in $\mathcal{L}_{CE-ReNode}$. Table \ref{ablation} presents the accuracy results for a single split of the Scene dataset in this ablation experiment.
The results show that each component contributes to the improvement in accuracy, with the combination of all components resulting in the most significant improvement.

By comparing the first row with the next three rows, we can observe the individual effects of each term on accuracy. Adding $\mathcal{L}_{CE-pseudo}$ has the greatest impact on accuracy improvement (as seen by comparing row 1 and row 3), while the addition of $\mathcal{L}_{smooth}$ has the smallest impact (as seen by comparing row 1 and row 4).

However, this does not imply that $\mathcal{L}_{smooth}$ does not play a role in enhancing our model. The three loss terms - $\mathcal{L}_{smooth}$, $\mathcal{L}_{CE-pseudo}$, and the weight $w_i$ in $\mathcal{L}_{CE-ReNode}$ - influence each other, and their contributions to the model are not linearly related. Instead, they interact with one another to collectively contribute significantly to the model's performance.

\subsubsection{\textcolor{black}{Scheduling strategies of $w_p$}}

\begin{table}[htbp]
\caption{\textcolor{black}{Scheduling strategies of $w_p$ on the Youtube dataset.}}
\label{wpFunction}
\centering
\begin{tabular}{llccc}
\hline
    &     Temporal schedule of $w_p$  &    & Accuracy \% \\\hline     
  1&    Linear  \( w_p = \frac{\text{epoch} - 1}{\text{Max}} \)      &      &  {\bf 56.75 }     \\
  2&    Exponential \( w_p = \exp\left(\frac{\text{epoch} - 1}{\text{Max}}\right) - 1 \)                     &      &   56.06     \\ 
  3&    Square root   \( w_p = \sqrt{\frac{\text{epoch} - 1}{\text{Max}}} \)        &     &   55.99     \\  
  4&    Square  \( w_p = \left(\frac{\text{epoch} - 1}{\text{Max}}\right)^2 \)       &     &   54.81     \\  
\hline
\end{tabular}
\end{table}

\textcolor{black}{The variable \( w_p \) serves to control the pseudo-label cross-entropy loss by initially assigning it a lower weight during the early phases of training, progressively increasing this weight as the number of epochs grows. This adjustment can be implemented through various monotonically increasing functions, including linear, square root, square, and exponential functions.}

\textcolor{black}{One approach is the linear function, defined as \( w_p = \frac{\text{epoch} - 1}{\text{Max}} \). This function starts at \( w_p = 0 \) when \( \text{epoch} = 1 \) and increases steadily to reach a value of $\approx$ 1 by \( \text{epoch} = 2000 \).}

\textcolor{black}{In addition to the linear function, the square root function, represented as \( w_p = \sqrt{\frac{\text{epoch} - 1}{\text{Max}}} \), and the square function \( w_p = \left(\frac{\text{epoch} - 1}{\text{Max}}\right)^2 \) offer different growth behaviors. The square root function begins with a rapid increase that eventually slows down, while the square function starts more gradually but accelerates as training progresses.
In contrast, the exponential function \( w_p = \exp\left(\frac{\text{epoch} - 1}{\text{Max}}\right) - 1 \) demonstrates a more pronounced acceleration, growing significantly faster as the training advances.}

\textcolor{black}{Table \ref{wpFunction} presents the classification accuracy achieved on the YouTube dataset when \( w_p \) follows four different functions. The results suggest that the model achieves the highest classification accuracy of 56.75\% when \( w_p \) is governed by the linear function. This steady and balanced increase in pseudo-label weight afforded by the linear function appears to be optimal, contributing to the improved accuracy observed in the model's performance.}

\subsubsection{Replace pseudo labels with ground-truth labels in $\mathcal{L}_{CE-pseudo}$}

In the loss term $\mathcal{L}_{CE-pseudo}$, we used the prediction matrix from the previous epoch as the pseudo label $\Yvect^p$, resulting in a more powerful cross-entropy loss term. For this experiment, we replaced the pseudo-labels $\Yvect^p$ with the ground-truth labels $\Yvect$. Of course, in real-world applications, the ground-truth labels of unlabeled samples are unknown, so we cannot use them in the loss function. However,  since this is a comparative experiment, we use the ground-truth labels in $\mathcal{L}_{CE-pseudo}$ to observe the potential effect that accurate pseudo-labels could achieve.

\begin{table}[]
\caption{Accuracy of replace pseudo labels with ground-truth labels in $\mathcal{L}_{CE-pseudo}$.}
\label{Y-YP}
\centering
\begin{tabular}{lccccc}
\hline
      & ORL           & Scene       & ALOI         &Youtube             & Handwritten  \\ \hline
 $\Yvect^p$ & 97.50 & 79.82  &  99.41 & 56.75 & 97.55   \\ 
 $\Yvect$ & 98.00 & 79.86 & 99.56 & 57.19  & 97.55   \\
 \hline

\end{tabular}
\end{table}

The first row of Table \ref{Y-YP} shows the accuracy of our RSGSLM on a single split across five datasets, while the second row shows the accuracy when pseudo labels in $\mathcal{L}_{CE-pseudo}$ are replaced with ground-truth labels for the same split. The values in the second row represent the upper limit of accuracy that pseudo-labels could help the model achieve, which corresponds to the accuracy obtained by using perfectly accurate pseudo-labels. We can see that improving the accuracy of pseudo labels can indeed enhance model performance. As shown in the table, the difference in performance is not substantial, indicating that our model has already utilized pseudo-labels effectively. Based on the earlier analysis of the contribution of the pseudo-label loss term $\mathcal{L}_{CE-pseudo}$  to model accuracy, we can conclude that our use of pseudo-labels is highly efficient.

\subsection{Computational complexity analysis}

In this section, we will analyze the computational complexity of our proposed method. For the two modules—Graph Learning and Re-weight Nodes — both utilize shallow learning techniques related to linear transformations, resulting in minimal computational costs. The computational complexity for these modules is $\mathcal{O}(n^3)$, where $n$ represents the number of samples. Our focus will be on the computational complexity of the GCN module, which is a deep learning approach, incurring relatively higher time and space costs.

Compared to other multi-view GCN methods, our approach employs only a single GCN, significantly reducing both time and space consumption relative to traditional GCN methods. Tables \ref{costgcn} and \ref{costgcnour} present a comparison of the matrix sizes used in training between the traditional GCN methods and our approach. In the graph learning process described in Section 4.1, we transform the original feature matrix $\Xvect^v\in \mathbb{R}^{n\times d_v}$ to $\Fvect^v  \in \mathbb{R} ^{n\times c}$. Without this conversion, the input to the GCN would be $ \Xvect_\ast = [ \Xvect^1, \ \Xvect^2, \ldots, \Xvect^V]  \in \mathbb{R}^{n \times (d_1 +d_2+ ...+d_V)}$. After our transformation, the input to the GCN becomes $ \Fvect_\ast = [ \Fvect^1, \ \Fvect^2, \ldots, \Fvect^V]  \in \mathbb{R}^{n \times (c \times V)}$. Using the Scene dataset as an example, we can observe from the last row of both tables that our conversion from $\Xvect\ast$ to $\Fvect\ast$ significantly reduces the size of multiple matrices during GCN training, thereby decreasing the space and time required for training.

Regarding time complexity, the first layer of each iteration in the traditional GCN has a time complexity of $\mathcal{O}(n^2d+n \cdot d \cdot d_h)$, where $d=(d_1 +d_2+ ...+d_V)$. The time complexity for the second layer is $\mathcal{O}(n^2d_h+n \cdot d_h \cdot c)$. Consequently, the total time complexity for each iteration of the traditional GCN is  $\mathcal{O}([n^2((d_1 +d_2+ ...+d_V)+d_h)+n \cdot ((d_1 +d_2+ ...+d_V) \cdot d_h+d_h \cdot c)])$. In contrast, the time complexity for each iteration in our method is $\mathcal{O}( [n^2((c \times V)+d_h)+n \cdot ((c \times V) \cdot d_h+d_h \cdot c)])$. 

For the Scene dataset, the time complexity of each iteration in the traditional GCN method is $\mathcal{O}([2688^2\cdot ((512+432+256+48)+28)+2688 \cdot ((512+432+256+48) \cdot 28+28 \cdot 8)])=\mathcal{O}([2688^2\cdot 1276+2688 \cdot 35168]) \approx \mathcal{O}(9.3\times10^9)$, while the time complexity for each iteration of our GCN method is $\mathcal{O}([2688^2\cdot ((8\times4)+28)+2688 \cdot ((8\times4) \cdot 28+28 \cdot 8)])=\mathcal{O}([2688^2\cdot 60+2688 \cdot 1120])\approx \mathcal{O}(4.3\times10^8)$. The difference between the two is over 20 times.

Our method significantly reduces both space and time consumption. This advantage is especially pronounced for datasets with larger dimensions (larger $d$), where our approach can save even more space and time compared to traditional methods.

\textcolor{black}{Table \ref{tcost} displays the running times of the methods GCN-multi, IMvGCN, and our proposed RSGSLM across all datasets (on the machine with I7-9700K CPU, Nvidia RTX 2080TI GPU, and 64G RAM).
The baseline model, GCN-multi, exhibits slow performance because it requires the execution of $V$ GCNs, a common issue associated with multi-view methods that necessitate running multiple GCNs. In contrast, our method operates efficiently as it requires only a single GCN, and the input GCN matrix $\Fvect\ast$ is significantly reduced in dimension due to the linear transformation applied.}

\begin{table}[]
\caption{Matrix Dimensions During the Training of a Traditional Two-Layer GCN.}
\label{costgcn}
\centering
\resizebox{1.0\linewidth}{!}{
\begin{tabular}{llllll}
\hline
&$\Svect$ & $\Xvect\ast$ & $\Wvect ^{\left(0\right)}$ & $\Wvect ^{\left(1\right)}$ & $\Zvect$\\ \hline
size &$ \mathbb{R} ^{n\times n}$  & $\mathbb{R}^{n \times (d_1 +d_2+ ...+d_V)}$  &  $ \mathbb{R} ^{(d_1 +d_2+ ...+d_V)\times d_h}$ &  $ \mathbb{R} ^{d_h\times c}$ &  $ \mathbb{R} ^{n\times c}$\\
size for Scene & $ \mathbb{R} ^{2688\times 2688}$  & $\mathbb{R}^{2688 \times (512+432+256+48)}=\mathbb{R}^{2688 \times 1248}$  &  $ \mathbb{R} ^{(512+432+256+48)\times 28}=\mathbb{R} ^{1248\times 28}$ &  $ \mathbb{R} ^{28\times 8}$ &  $ \mathbb{R} ^{2688\times 8}$\\

 \hline
\end{tabular}
}
\end{table}

\begin{table}[]
\caption{Matrix Dimensions During the Training of Two-Layer GCN in our Approach.}
\label{costgcnour}
\centering
\resizebox{0.8\linewidth}{!}{
\begin{tabular}{llllll}
\hline
&$\Svect$ & $\Fvect\ast$ & $\Wvect ^{\left(0\right)}$ & $\Wvect ^{\left(1\right)}$ & $\Zvect$\\ \hline
size &$ \mathbb{R} ^{n\times n}$  & $\mathbb{R}^{n \times (c \times V)}$  &  $ \mathbb{R} ^{(c \times V)\times d_h}$ &  $ \mathbb{R} ^{d_h\times c}$ &  $ \mathbb{R} ^{n\times c}$\\
size for Scene & $ \mathbb{R} ^{2688\times 2688}$  & $\mathbb{R}^{2688 \times (8\times4)}=\mathbb{R}^{2688 \times 32}$  &  $ \mathbb{R} ^{(8\times4)\times 28}=\mathbb{R} ^{32\times 28}$ &  $ \mathbb{R} ^{28\times 8}$ &  $ \mathbb{R} ^{2688\times 8}$\\

 \hline
\end{tabular}
}
\end{table}

\begin{table}[]
\caption{\textcolor{black}{Running time in seconds for GCN-Multi, IMvGCN, and RSGSLM.
}}
\label{tcost}
\centering
\resizebox{0.5\linewidth}{!}{
\begin{tabular}{lrrr}
\hline
Dataset     & GCN-multi & IMvGCN & RSGSLM\\ \hline
ALOI        &   95.11   & 7.79      &  2.97      \\
ORL            & 247.37 &  37.2       &  11.28    \\
Scene       &    812.22 &  114.03       &  45.55     \\
Handwritten      & 363.56 &  134.84    & 41.60
\\
Youtube &     951.14 &  112.78   & 18.31 \\ 
MNIST         &6449.22 &  639.43      & 140.33\\ 
\hline
\end{tabular}
}
\end{table}

\section{Conclusion}
%%%%%%%%%%%%%%%%%%%%%%%
In this paper, we propose a Re-node Self-taught Graph-based Semi-supervised Learning for Multi-view data (RSGSLM), specifically designed to handle non-graph data from multiple views.  Our method uses linear projection to create low-dimensional features and constructs a fused graph, which is subsequently input into one GCN. We then re-weight the labeled samples and train the GCN for semi-supervised classification. Throughout training, we consider several loss functions, such as a re-weighted cross-entropy loss, a pseudo-label loss, and a manifold regularization loss. These three loss functions interact with each other to collectively improve the model's performance. The entire model is trained within a single, compact GCN framework.

We integrate pseudo-labels into the GCN training loss function, allowing the network to benefit from pseudo-labels through this loss term. This pseudo-label loss does not introduce any additional iterations or parameters; it considers all prediction results as pseudo-labels and adjusts their weights based on the confidence level. By using a temporal global weight, we address the issue of inaccurate pseudo-labels in the early training phase, minimizing their potential negative impact on the model.

\textcolor{black}{In real-world applications, the proposed model offers an efficient and robust framework for processing multi-view image data, particularly in environments where labeled data are scarce. Its ability to learn from limited supervision makes it highly suitable for domains where manual annotation is costly, time-consuming, or impractical. This framework has the potential to support several critical sectors that drive national economic and social development. Notable examples include hydrological monitoring within smart water management systems, where accurate analysis of multi-source imagery can aid in flood prediction and water resource optimization; video surveillance in smart cities, where multi-camera data fusion enhances object tracking, anomaly detection, and public safety; and environmental perception for autonomous driving, where integrating views from multiple sensors improves scene understanding and decision-making in complex traffic environments. Beyond these examples, the model can be extended to other multi-view visual intelligence tasks—such as precision agriculture, infrastructure inspection, and ecological monitoring—where robust performance under limited supervision is essential for real-world deployment.}

Nevertheless, there are several areas where future research could further enhance the capabilities and applications of GCNs. Here are some important directions for future work:
\textbf{Inductive Learning Models}: We aim to develop inductive models that can generalize to unseen samples, enabling the inference of labels on new, previously unseen data.
\textbf{Transfer Learning}: This approach is particularly useful when there is limited labeled data for the target task but abundant labeled data for related tasks. Developing transfer learning methods for GCNs could facilitate knowledge transfer from one graph domain to another, thereby improving performance in scenarios with limited labeled data.
\textbf{Attention Mechanisms}: These mechanisms mimic the human cognitive ability to focus on specific elements when processing complex information. Incorporating more advanced attention mechanisms into GCNs could help capture the importance of different nodes and edges more effectively, enhancing performance on complex tasks.

\end{document}